\documentclass[10pt,twocolumn,letterpaper]{article}

\usepackage{wacv}
\usepackage{times}
\usepackage{epsfig}
\usepackage{graphicx}
\usepackage{amsmath}
\usepackage{amssymb}
\usepackage{booktabs}
\usepackage{caption}
\usepackage{subcaption}
\usepackage[accsupp]{axessibility}

\def\wacvPaperID{990} 

\wacvalgorithmstrack   

\wacvfinalcopy 

\ifwacvfinal
\usepackage[breaklinks=true,bookmarks=false]{hyperref}
\else
\usepackage[pagebackref=true,breaklinks=true,colorlinks,bookmarks=false]{hyperref}
\fi

\begin{document}

\title{Unsupervised 4D LiDAR Moving Object Segmentation in Stationary Settings \\ with
Multivariate Occupancy Time Series}

\author{Thomas Kreutz \hspace{1cm} Max Mühlhäuser \hspace{1cm} Alejandro Sanchez Guinea\\
Telekooperation Lab, Technical University Darmstadt\\
{\tt\small \{kreutz, max, sanchez\}@tk.tu-darmstadt.de}}

\maketitle
\thispagestyle{empty}

\begin{abstract}
   
In this work, we address the problem of unsupervised moving object segmentation (MOS) in 4D LiDAR data recorded from a stationary sensor, where no ground truth annotations are involved. Deep learning-based state-of-the-art methods for LiDAR MOS strongly depend on annotated ground truth data, which is expensive to obtain and scarce in existence.
To close this gap in the stationary setting, we propose a novel 4D LiDAR representation based on multivariate time series that relaxes the problem of unsupervised MOS to a time series clustering problem. More specifically, we propose modeling the change in occupancy of a voxel by a multivariate occupancy time series (MOTS), which captures spatio-temporal occupancy changes on the voxel level and its surrounding neighborhood. To perform unsupervised MOS, we train a neural network in a self-supervised manner to encode MOTS into voxel-level feature representations, which can be partitioned by a clustering algorithm into moving or stationary. Experiments on stationary scenes from the Raw KITTI dataset show that our fully unsupervised approach achieves performance that is comparable to that of supervised state-of-the-art approaches. 

\end{abstract}


\section{Introduction}



Understanding an urban environment in terms of its moving or static entities is a crucial aspect for scene understanding~(\eg,~\cite{aygun20214d}), autonomous driving agents~(\eg,~\cite{postica2016robust, schreiber2020motion}), consistent mapping~(\eg,~\cite{chen2021moving}), pedestrian safety, and intelligent transportation systems in smart cities~(\eg,~\cite{muhlhauser2020street, harris2019mlk}).
In particular, LiDAR moving object segmentation (MOS) is a task to classify the points of a scene into being dynamic or static.

The research on end-to-end approaches for LiDAR object detection, semantic segmentation, instance segmentation, and panoptic segmentation has matured over the past years~\cite{guo2020deep} and large-scale autonomous driving datasets like SemanticKITTI~\cite{geiger2012cvpr, behley2019iccv}, NuScenes~\cite{nuscenes2020, nuscenes2022panoptic}, or Waymo~\cite{Sun_2020_CVPR, Ettinger_2021_ICCV} have been the essential ingredient for developing state-of-the-art approaches. Unfortunately, annotated data for LiDAR MOS is scarce~\cite{chen2022automatic}. Recently, an annotated MOS benchmark dataset based on SemanticKITTI has been proposed in~\cite{chen2021moving}, which fostered promising research about end-to-end approaches for MOS in the autonomous driving setting~(\eg,~\cite{mohapatra2021limoseg, mersch2022receding, gu2022semantics}). However, the lack of annotated datasets limits the practical application of supervised end-to-end MOS deep learning models to scenarios where data has not been recorded with the same sensor setup~\cite{chen2022automatic}.



A potential solution to the described issue is unsupervised methods because they do not depend on annotated data and generalize better to arbitrary data distributions~\cite{bevsic2022unsupervised, yuan2019unsupervised}. For instance, self-supervised scene flow methods can be used for unsupervised MOS, but their performance is inferior to state-of-the-art supervised methods~\cite{mersch2022receding}. 

In contrast to previous work, we propose a fully unsupervised 4D LiDAR MOS approach that generalizes to data recorded from arbitrary stationary LiDAR sensors, and achieves results that are comparable to that of supervised state-of-the-art approaches. 
Previous work has shown that movement appears with occupancy change patterns in occupancy time series~\cite{engel2018deep}. On this basis, it can be hypothesized that multivariate occupancy time series (MOTS) are an effective data modality to identify motion in spatio-temporal neighborhoods of point cloud videos. In our paper, we answer the following hypothesis: \textit{Multivariate time series are an effective data-modality for unsupervised MOS in stationary LiDAR point cloud videos.} 

We propose MOTS as a novel 4D LiDAR representation that allows using self-supervised representation learning to distinguish between moving and static parts in a stationary LiDAR scene. More specifically, a voxel is represented by a MOTS that effectively models spatio-temporal occupancy changes of the voxel and its surrounding neighborhood. Following recent advances in self-supervised learning for multivariate time series (\eg, \cite{franceschi2019unsupervised, tonekaboni2021unsupervised}), we first encode MOTS in short time windows with a neural network to a spatio-temporal voxel embedding. Afterward, we cluster the resulting embeddings of each voxel for unsupervised MOS. Therefore, our approach relaxes MOS to a multivariate time series clustering problem. 

We show the effectiveness of MOTS for unsupervised MOS by quantitative evaluations on publicly available stationary data from the Raw KITTI dataset~\cite{geiger2012cvpr} and a qualitative evaluation on stationary data we recorded with a Velodyne VLP-16 sensor. 
Our main contributions are:
\begin{itemize}
    \item A novel representation of 4D point clouds for representation learning of spatio-temporal occupancy changes in a local neighborhood of stationary LiDAR point cloud videos, which we call MOTS
    \item An unsupervised MOS approach for stationary 4D LiDAR point cloud videos based on MOTS
\end{itemize}

\section{Related Work}
The majority of closely related work on moving object segmentation (MOS) can be categorized into dynamic occupancy grid mapping (\eg, \cite{nuss2018random, schreiber2021dynamic}), scene flow (\eg, \cite{liu2019flownet3d, baur2021slim}), and moving object segmentation methods (\eg, \cite{chen2021moving, mersch2022receding}).


\subsection{Dynamic Occupancy Grid Mapping}
Occupancy grid mapping estimates probabilities for the occupancy of grid cells. Furthermore, dynamic occupancy grid mapping (DOGMA) aims to learn a state vector for each grid cell that consists of occupancy probability and velocity~\cite{nuss2018random}.
An effective dynamic occupancy grid mapping based on finite random sets has been proposed in~\cite{nuss2018random}. 

Using the result in~\cite{nuss2018random} as a basis, various deep learning-based methods that learn DOGMAs have been proposed. For instance, the work in~\cite{engel2018deep} uses the DOGMA from~\cite{nuss2018random} as an input to a neural network that learns to predict bounding boxes for moving objects. The work in~\cite{schreiber2020motion} learns DOGMAs to estimate motion of objects in the scene with a neural network in a stationary setting. They use the DOGMA obtained from the approach in~\cite{nuss2018random} as a basis to train their model end-to-end, and their work was extended in~\cite{schreiber2021dynamic} to the non-stationary setting.

In spite of their success, the described methods depend on a DOGMA to find moving objects, and they are limited to 2D birds-eye view (BEV) maps. Today, in other related tasks such as semantic segmentation, projection based deep learning methods are getting outperformed by methods that operate directly in the 3D or 4D domain~\cite{zhou2020cylinder3d}. In contrast, our method is designed for raw 4D point clouds and does not depend on occupancy grid mapping methods.

\subsection{Scene Flow}

Scene flow methods learn a displacement vector for any point in frame $t$ to frame $t+1$. Hence, a scene flow method can be extended to a MOS approach. For instance, clustering point positions together with their corresponding scene flow vectors have been used in~\cite{liu2019flownet3d} to obtain an unsupervised motion segmentation. Furthermore, the work in~\cite{baur2021slim} showed that scene flow based on a self-supervised method can learn to segment motion as a byproduct.  However, the downside of scene flow methods is that (a) there is no clear correspondence between points across frames in noisy point clouds and (b) only using two frames might not contain enough information for all moving points in the scene, particularly when dealing with slow moving objects, as outlined in~\cite{mersch2022receding}. These limitations can explain the inferior results for scene flow-based MOS on the SemanticKITTI MOS benchmark~\cite{chen2021moving, mersch2022receding}.
In comparison, our approach can learn motion from a larger temporal context by including more than two frames. Furthermore, a trivial correspondence between voxels across all frames exists in our approach because it is designed on the voxel level.

\subsection{Motion/Moving Object Segmentation}

Recently, a benchmark and a supervised model (LMNet) based on range images for MOS has been proposed in~\cite{chen2021moving}. The authors extended their work with an automatic labeling approach that is based on a map cleaning method in~\cite{chen2022automatic}, which makes it more robust to unseen environments and improves the performance. The work in~\cite{mohapatra2021limoseg} proposed a method based on BEV which is faster than LMNet but with inferior segmentation performance.

A method for 4D MOS has been proposed recently in~\cite{mersch2022receding}, where predictions are made from a 4D volume of the point cloud video. Further, a Bayesian filter taking previous predictions into account is proposed to filter out noise. The model in~\cite{mersch2022receding} makes use of sparse convolutions~\cite{choy20194d}, which achieve better performance than projecting the point cloud to a two-dimensional range image representation.

Fusing semantic predictions with moving object predictions has been shown to increase the performance in~\cite{chen2021moving}. Using semantic features for MOS has been used specifically in~\cite{gu2022semantics}. In this case, the semantic features are learned individually on each frame and the moving object segmentation mask is learned afterward jointly from sequences of the resulting semantic features and range images.

All the aforementioned approaches show a promising performance, but they rely on annotations to train their approach. In comparison, we are the first to propose an unsupervised approach for MOS in the stationary setting which does not depend on occupancy grid mapping, map cleaning or scene flow methods. At the same time, our approach is based on multiple frames.



\begin{figure*}[t]
    \centering
    \includegraphics[width=\linewidth]{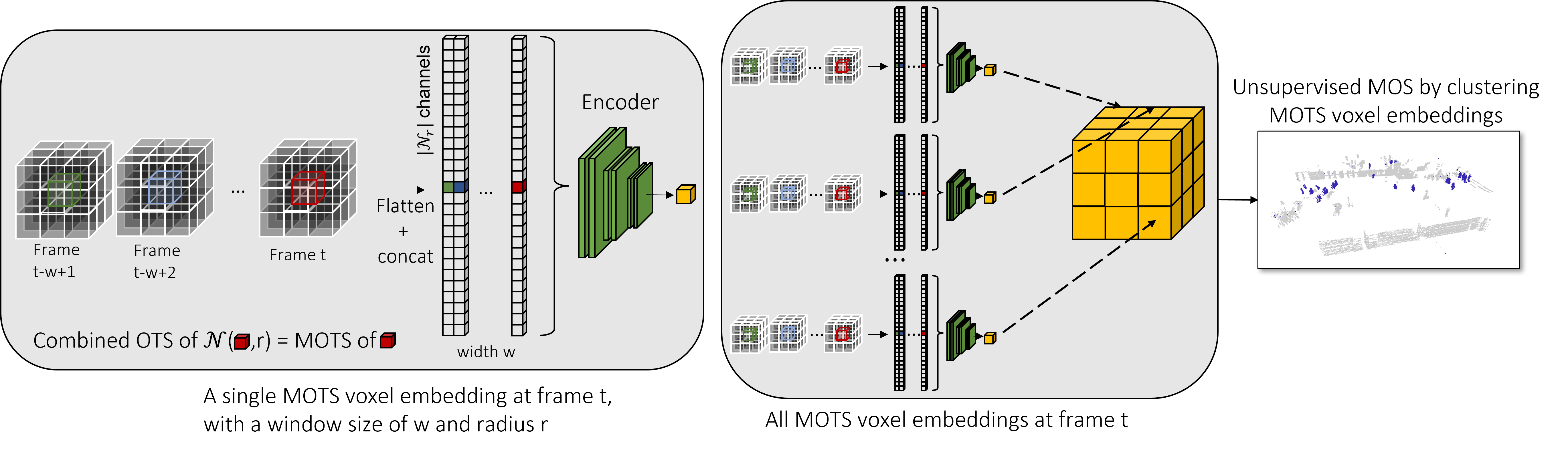}
    \caption{Overview of the proposed approach}
    \label{fig:overview}
\end{figure*}

\subsection{Unsupervised Segmentation of Time Series}
Recent advances in self-supervised multivariate time series representation learning (\eg, ~\cite{franceschi2019unsupervised, luxem2020identifying, tonekaboni2021unsupervised}) 
have shown that different states of a system (measured by a multivariate time series) can be learned for each time step in a self-supervised manner. The learned representations at each time step can then be clustered to obtain an unsupervised segmentation of the time series. 

To the best of our knowledge, our work is the first to adapt this idea to the point cloud domain. We consider occupancy states at each point in time of a voxel as discrete time series measurements and exploit the dependency between occupancy changes in the spatio-temporal neighborhood of a voxel for unsupervised MOS.

\section{Approach}

\subsection{Problem setup}
Given a point cloud video recorded from a stationary LiDAR sensor, our goal is to produce an unsupervised segmentation of the scene into moving and stationary points without having to rely on annotated data. More specifically, the goal is to perform unsupervised moving object segmentation (MOS) solely on raw, stationary LiDAR point cloud videos. This problem is of practical use in smart cities where LiDAR sensors can be mounted on, for instance, street lamps that cover a large area of the city~\cite{muhlhauser2020street} and moving objects have to be identified. Another crucial use case is identifying moving objects in the traffic around a stationary autonomous vehicle that waits to drive onto a busy road. 

\subsection{Overview}
We propose a novel representation of point cloud videos in order to learn spatio-temporal representations of single voxel cells. The proposed representation relaxes unsupervised MOS to a multivariate time series clustering problem. 


Figure~\ref{fig:overview} summarizes our method. At frame $t$, we compute occupancy time series (OTS) of length $w$ for all voxel cells in the frame. Given a spatial radius $r$, we construct multivariate occupancy time series (MOTS) from the OTS of a voxel and all OTS in its surrounding neighborhood. Each channel of MOTS effectively captures the occupancy change in local spatio-temporal neighborhoods of the scene. We assume that movement appears similarly across MOTS from different voxels and clustering MOTS separates moving voxels from stationary voxels. Hence, with MOTS the MOS problem is relaxed to a multivariate time series clustering problem. 

Given a MOTS point cloud video representation at any frame $t$, a neural network encodes all MOTS in frame $t$ to a feature representation that can distinguish moving from stationary voxel states. This kind of representation learns and encodes spatio-temporal occupancy changes such that a clustering algorithm can perform unsupervised MOS.

\subsection{Multivariate Occupancy Time Series}


Our approach is designed for voxelized point cloud videos. A voxelgrid is a set of voxels \mbox{$V \subseteq \mathbb{R}^{w/m \times h/m \times d/m}$}, with a grid resolution of $m$, height $h$, width $w$, and depth $d$. An ordered sequence of 3D voxelgrids $VS \in (V_{1}, ..., V_{N})$ can be considered a video, with $N$ being the number of frames. A voxel $v \in V$ can have one of two states: occupied or free. The state of a voxel at time $t$ is modelled by a function \hbox{$S : V \times \mathbb{N} \longrightarrow \mathbb{B}$}, with the interpretation that $0=\text{free}$ and $1=\text{occupied}$.  Assuming data recorded from a stationary LiDAR, there is a bidirectional mapping from any voxel $v \in V_{k}$ to a voxel $v' \in V_{l}$, $k \neq l$, such that $v == v'$. As a result, at any point in time $t$, for any voxel $v_{i}$, we can define its occupancy time series $OTS_{i,t} \in \mathbb{B}^{w} $ as
\begin{equation}
    OTS_{i, t} = [S(v_{i}, t-(w-1)), ..., S(v_{i}, t-1), S(v_{i}, t)] 
\end{equation}
with $w$ being the time series length, and $S(v_{i},\cdot)$ measuring the occupancy of $v_{i}$ at each point in time.

We define MOTS as a multivariate collection of OTS, where we consider the OTS of voxels $v_{j}$ in a spatial neighborhood around $v_{i}$ as additional channels. Given a spatial radius of $r$ around $v_{i}$ with a voxel grid resolution of $m$ units in an arbitrary euclidean space, we define a set \hbox{$R = \{-r, -r+m, ...,  0, ... , r-m, r\}$} that includes all possible discrete distances within radius $r$ around 0 as the center. We then compute a neighborhood distance matrix $\mathcal{N}_{r} = R \times R \times R$ with a 3-fold Cartesian product over $R$ by considering each element in $\mathcal{N}_{r}$ as a row. 
$\mathcal{N}_{r}$ holds the distances to all reachable voxels within radius $r$ considering an arbitrary voxel $v_{i}$ as the center. An element-wise addition of each row in $\mathcal{N}_{r}$ with $v_{i}$ computes the neighborhood
\begin{equation}
    \mathcal{N}(v_{i}, r) = \mathcal{N}_{r} + v_{i}
\end{equation}

with $\mathcal{N}_{r} + v_{i}$ being the shorthand notation of adding $v_{i}$ to each row of $\mathcal{N}_{r}$ ~(\cite{Goodfellow-et-al-2016}). 

Given the neighborhood $\mathcal{N}(v_{i}, r)$ of $v_{i}$ with radius $r$, we define a multivariate occupancy time series \hbox{$MOTS_{i, t} \in \mathbb{B}^{|N (v_{i}, r)| \times w}$ of $v_{i}$} as 

\begin{equation}
    MOTS_{i, t} = \{ OTS_{j, t} \ | \ v_{j} \in \mathcal{N}(v_{i}, r)\}
\end{equation}
where the channels of $MOTS_{i,t}$ are composed by the OTS of each $v_{j} \in \mathcal{N}(v_{i}, r)$.

\paragraph{MOTS for Non-Stationary LiDAR.}
While the focus of our work is on the stationary case, MOTS can also be computed in the non-stationary case. For a non-stationary LiDAR, we assume to have the pose information given by, \eg, a SLAM approach~\cite{geiger2012cvpr}. Given the poses, we transform each frame to the pose of the first frame to again obtain a bidirectional mapping from any voxel $v \in V_{k}$ to $v' \in V_{l}$, $k \neq l$, such that $v == v'$. The latter property allows computing MOTS for each voxel in a non-stationary setting.

\subsection{Efficiently Transforming 4D Point Clouds into MOTS}
A dense MOTS representation of 4D point clouds is inefficient because most of the space in all frames is empty. Hence, following related work~(\eg, \cite{choy20194d}) we adopt a sparse tensor representation and only store/compute MOTS of voxels that are occupied. We can represent each frame $V_{t}$ with $0 < t \leq N$ of a 4D point cloud as a sparse tensor, which we consider a pair ($\mathcal{V}^{t}_{sparse}$, $\mathcal{F}^{t}_{sparse}$) consisting of a set of voxels $\mathcal{V}^{t}_{sparse}$ that we define as

\begin{equation}
    \mathcal{V}^{t}_{sparse} = \{ (v_{i}, t) \ | \  S(v_{i}, t)=1 \wedge v_{i} \in V_{t}\} 
\end{equation}

and its corresponding set of MOTS features $\mathcal{F}^{t}_{sparse}$ that we define as
\begin{equation}
    \mathcal{F}^{t}_{sparse} = \{ MOTS_{i, t} \  | \  S(v_{i}, t)=1 \wedge v_{i} \in V_{t} \} 
\end{equation}

In practice, we make use of vectorized operations and a performant parallel hashmap Python implementation\footnote{https://github.com/atom-moyer/getpy} to efficiently compute the MOTS features for each voxel. 



\subsection{Unsupervised Moving Object Segmentation with MOTS}
We relax unsupervised MOS with MOTS to a time series clustering problem. Due to the enormous amount of available training data and high dimensionality of the time series, we leverage deep learning to learn the underlying structure of the data and use an autoencoder (AE) as a feature extractor. More specifically, we use an AE based on 1D convolutions to learn feature representations of MOTS. 


The AE consists of an encoder and a decoder part. The encoder $f: \mathbb{R}^{d} \mapsto \mathbb{R}^{e}$ maps a $d$-dimensional input data point $x \in \mathbb{R}^{d}$ to an $e$-dimensional (latent) code representation $z \in \mathbb{R}^{e}$. The decoder is a function $f: \mathbb{R}^{e} \mapsto \mathbb{R}^{d}$ that maps the $e$-dimensional code vector $z$ back to a $d$-dimensional output $\hat{x}$, with the goal to be as similar as possible to the input $x$, i.e., $g(f(x)) = \hat{x} \approx x$ with $f(x) = z$ and  $g(z) = \hat{x}$. To this end, the AE is trained using the well-known mean squared error (MSE) loss function
\begin{equation}
    \mathcal{L} = MSE(x, \hat{x})
\end{equation}
in order to minimize the reconstruction error.

\paragraph{Unsupervised MOS.}
For each frame in $VS$, we encode the MOTS of all occupied voxels with the encoder $f$. Afterward, a clustering model partitions the voxel embeddings into a moving or stationary state. In this work, we perform unsupervised MOS by clustering the voxel embeddings with a gaussian mixture model (GMM). We empirically found the GMM to significantly perform better than, \eg, $k$-means. 

\begin{figure}[t]
    \centering
    \includegraphics[width=0.9\linewidth]{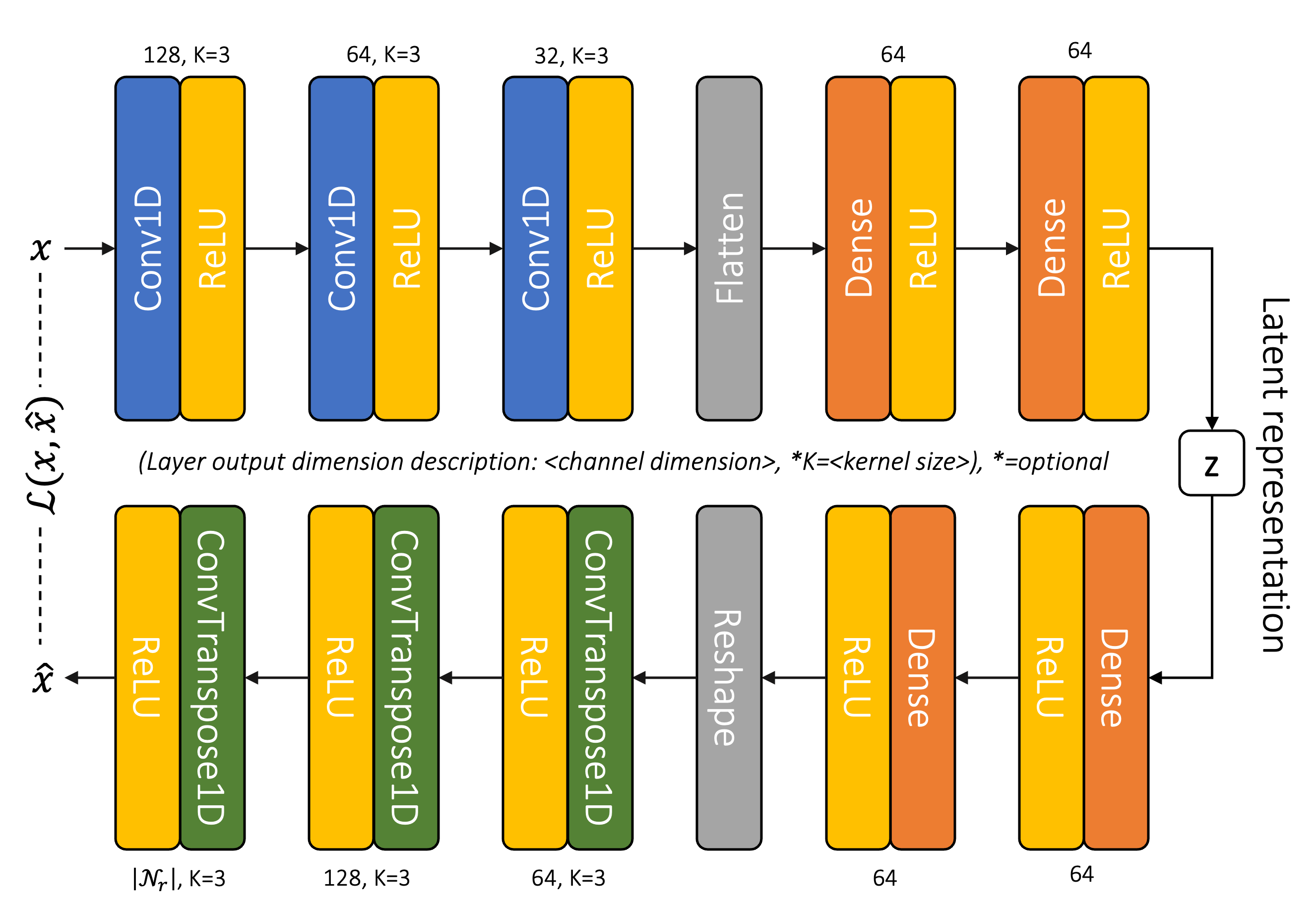}
    \caption{Architecture of our 1D CNN Autoencoder}
    \label{fig:autoencoder}
\end{figure}

\subsection{Architecture}

We depict the architecture of the AE we use in this work in Figure~\ref{fig:autoencoder}. The encoder is composed of three 1D convolutional layers, each having a kernel size of three. Afterward, we use two fully connected (FC) dense layers to project the output of the last convolutional layer to the $e$-dimensional code vector. The decoder consists of the reversed FC dense layers and three 1D transposed convolutions to reconstruct the input from the code vector. After each layer, we use the ReLU activation function as the non-linearity. \textit{We use this straightforward baseline model to highlight the effectiveness of MOTS in distinguishing local occupancy changes.}


\section{Evaluation}

\begin{table*}[t]
    \centering
    \begin{tabular}{cccccc}
          \multicolumn{2}{c}{} & \multicolumn{3}{c}{$mIoU (\uparrow$)} & \\ \toprule
         Setting   & Approach & City 1 & City 2 & Campus 1 & avg \\
        \midrule
              
    Supervised &    LMNet~\cite{chen2021moving}          &  .281    &   .557    &  .787  & .541 \\
    Supervised   & 4DMOS w/o BF~\cite{mersch2022receding}   &  .639  & .743 & .940  & .774 \\ 
    Supervised   & 4DMOS w/ BF p=0.25~\cite{mersch2022receding}   &  .567  &  .660 &  .944  & .723 \\
    Supervised   & 4DMOS w/ BF p=0.5~\cite{mersch2022receding}   &  .630  &  .769 &  .946  & .781 \\ 
         \midrule
   
     
     \textbf{Unsupervised}   &  \textbf{Ours $r=2, e=32, k=20, w=15$}  & \textbf{.792} & \textbf{.840} &  \textbf{.791}  &  \textbf{.808} \\
     \textbf{Unsupervised}   &  \textbf{Ours $r=2, e=32, k=20, w=20$}  & \textbf{.806 } & \textbf{.839} &  \textbf{.792}  & \textbf{ .812} \\
        \bottomrule
    \end{tabular}
    \caption{Summary of the mIoU results on stationary KITTI scenes against the state-of-the-art}
    \label{tab:miou_raw_kitti}
\end{table*}

In this section, we first describe our experimental setup and design. Afterward, we quantitatively evaluate our approach against supervised state-of-the-art approaches for MOS on stationary scenes from the KITTI dataset. In addition, we investigate the influence of hyperparameters on the overall performance of our approach. Finally, we perform a qualitative evaluation with Velodyne VLP-16 LiDAR data. Our source code and data is publicly available on github \url{github.com/thkreutz/umosmots}.


\subsection{Dataset and Metric}

\paragraph{Raw KITTI~\cite{geiger2012cvpr}.} 
To the best of our knowledge, a large-scale dataset with point-wise moving object annotations for stationary LiDAR sensors is not publicly available. The SemanticKITTI MOS dataset~\cite{behley2019iccv, chen2021moving} includes annotations for moving objects. However, there are not enough stationary frames in the validation sequence (see Supplementary Section 1) for a strong evaluation, and the annotations for the test sequences are not publicly available. 

For a meaningful evaluation against the state-of-the-art in a stationary setting, we manually annotated sequences from the Raw KITTI dataset~\cite{geiger2012cvpr}. The data in Raw KITTI has been recorded with a 64-beam Velodyne HDL-64E LiDAR sensor at a 10Hz framerate. We manually annotated three stationary scenes from the ``Campus'' and ``City'' categories summarized in Table~\ref{tab:annotated_kitti}. These three scenes have in total 378 frames for evaluation. 

For a fair comparison against the state-of-the-art, our AE model is trained only on the training sequences \hbox{$\{i \ | \ 0 \leq i < 11\} \setminus \{ 8 \}$} of the SemanticKITTI dataset. One MOTS per voxel is one training example, which leads to an enormous amount of training data. For this reason, we decided to train only on the first 200 frames of each sequence.

\paragraph{Velodyne VLP-16.} For a qualitative evaluation on a different sensor, we recorded eight stationary LiDAR scenes with a 16-beam Velodyne VLP-16 LiDAR around the Campus of the TU Darmstadt. The data was collected at a framerate of 20Hz, which is twice the framerate of Raw KITTI. 


\begin{table}[t]
    \centering
    \begin{tabular}{ccc}
    \toprule
        Name & \#Frames \\
       \midrule
        2011\_09\_26\_drive\_0017\_sync (City 1)  & 114 \\
        2011\_09\_26\_drive\_0060\_sync (City 2)  & 78 \\
        2011\_09\_28\_drive\_0016\_sync (Campus 1) & 186 \\[5pt]
        Total & 378 \\
        \bottomrule
    \end{tabular}
    \caption{Overview of name, category, and number of frames for each annotated scene}
    \label{tab:annotated_kitti}
\end{table}




\paragraph{Metric.}
We quantify the performance of our approach against the state-of-the-art by following related work~\cite{chen2021moving} and use the intersection-over-union (IoU) metric. 
To evaluate the performance over all frames, we compute the mean of the $IoU$ from each frame, which is known as the mean-intersection-over-union (mIoU).

\subsection{Implementation Details}
We train our AE models with the Adam optimizer, \hbox{ $\text{batch size}=1024$ }, \hbox{$\text{learning rate} = 1e - 4$}, and embedding dimension \hbox{$e \in \{16, 32\}$} for two epochs, which takes around 8--24 hours on an RTX A4000 16GB GPU. We evaluate different window sizes $w\in \{8, 10, 15, 20\}$, where 8 is the best setting found in~\cite{chen2021moving} and 10 in~\cite{mescheder2019occupancy}. We further evaluate the neighborhood 
radius settings \hbox{$r=\{1,2\}$} for MOTS, which lead to possible MOTS of dimensions \hbox{$\{27,125\} \times \{8, 10, 15, 20\}$}.

Regarding clustering, the number of clusters for the GMM is evaluated between \hbox{$k=\{10,15,20\}$}. We train one GMM per scene on $200,000$ uniformly sampled embeddings from the first ten frames of each scene to speed up the training. The final predictions are obtained by computing the $IoU$ for each cluster to the ground truth ``moving'' class on the first frame of the respective sequence. We empirically find that our approach usually predicts moving voxels across $1$ and $3$ clusters, each having a ground truth overlap of at least $0.15$. Therefore, we automatically map all clusters with an $IoU$ overlap of at least $0.15$ to the class ``moving''. In practice, this overlap can be found with minimal effort by a domain expert, or a small scene can be annotated.


\subsection{Experimental Design}
We evaluate our approach against the state-of-the-art in a stationary setting of the Raw KITTI dataset. In particular, we evaluate against the two best recent supervised state-of-the-art methods for LiDAR MOS\footnote{At the time of writing this work (July 2022)}: 
LMNet~\cite{chen2021moving} and 4DMOS~\cite{mersch2022receding}. Both approaches have been trained on SemanticKITTI data. For this reason, a comparison through an experiment on the Raw KITTI data against the latter approaches can in fact be made because the sensor setup is equivalent. We argue that a model trained on data from sequences of a mostly moving sensor (with ego-motion compensation) should perform well on data recorded from a stationary vehicle. This situation occurs naturally while a vehicle is stopping at a red light or is waiting to merge into a busy road. To the best of our knowledge, a distinction between a moving or stationary ego-vehicle is not made in evaluations on large-scale autonomous driving datasets (\eg, SemanticKITTI, NuScenes~\cite{nuscenes2020}, Waymo~\cite{Sun_2020_CVPR}, Argoverse~\cite{Argoverse}). For this reason, we believe the results of our experiments are a valuable contribution to the community.

In the remainder of this evaluation, we follow related work~\cite{liu2019flownet3d, baur2021slim} and remove the ground. However, because we operate in a stationary setting, where the FOV and location of the sensor do not change, we remove the ground by simply thresholding the z-axis to $-1$. We furthermore evaluate our approach against the state-of-the-art on the voxel level. To this end, all point-wise predictions by LMNet~\cite{chen2021moving} and 4DMOS~\cite{mersch2022receding} are mapped to its respective voxel.




\subsection{Results on Raw KITTI}

\begin{figure*}[t]
    \centering
    \includegraphics[width=0.9\linewidth]{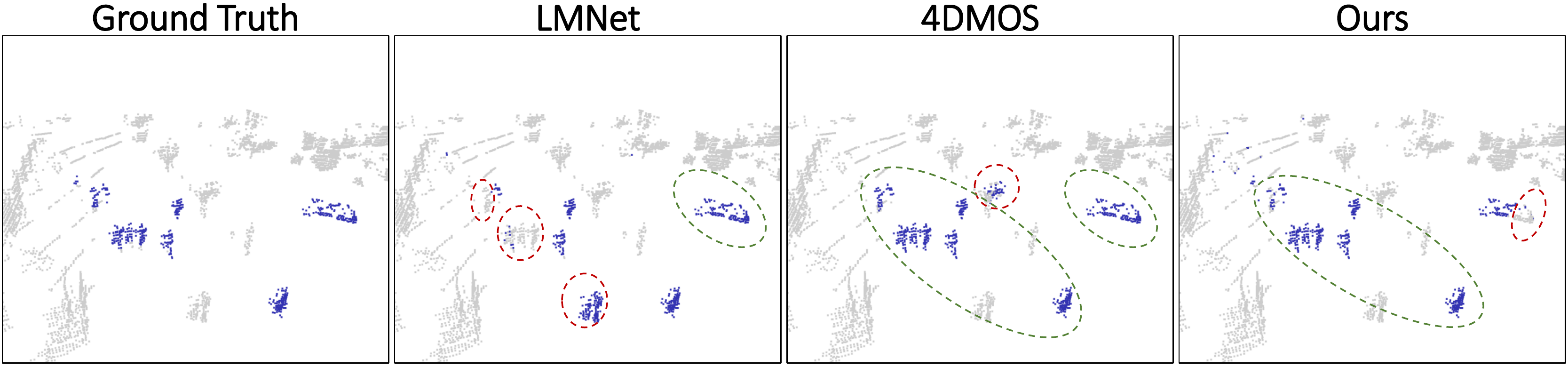}
    \caption{Qualitative comparison to the ground truth and state-of-the-art approaches on the Campus 1 scene}
    \label{fig:qualitative_campus}
\end{figure*}

\begin{figure*}[t]
    \centering
    \includegraphics[width=0.9\linewidth]{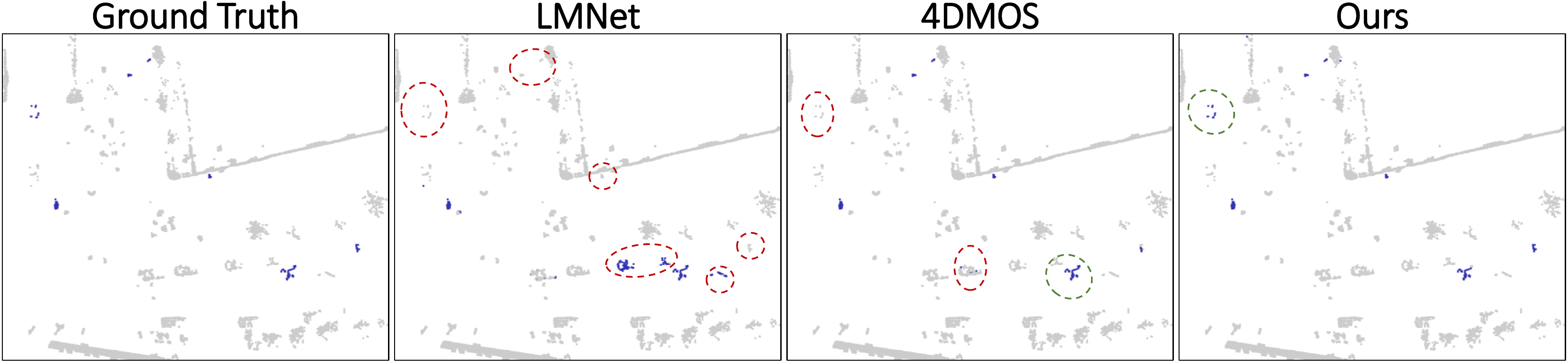}
    \caption{Qualitative comparison to the ground truth and state-of-the-art approaches on the City 2 scene}
    \label{fig:qualitative_city}
\end{figure*}

\begin{figure*}[t]
    \centering
    \includegraphics[width=0.8\linewidth]{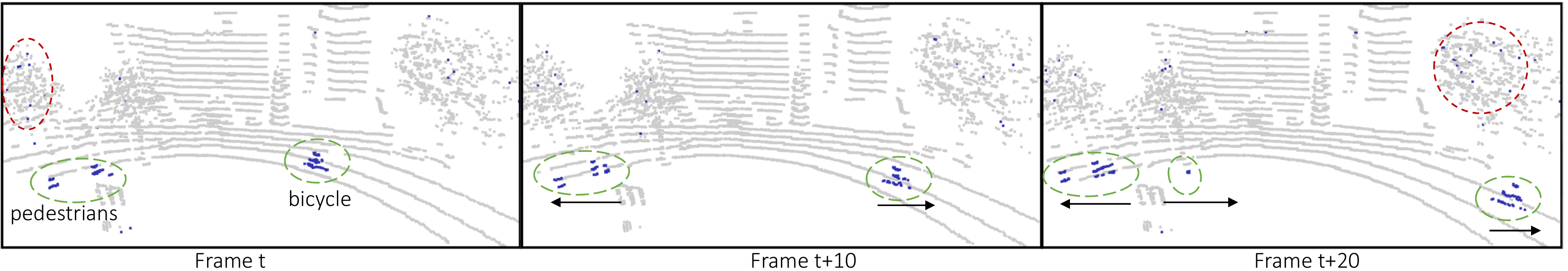}
    \caption{Qualitative results on data recorded from a Velodyne VLP-16 LiDAR}
    \label{fig:qualitative_vlp16}
\end{figure*}

We compare the two best-performing configurations of our approach against state-of-the-art MOS approaches on our three annotated scenes from the Raw KITTI dataset. The results in Table~\ref{tab:miou_raw_kitti} show that our unsupervised approach reaches comparable performance to the state-of-the-art w.r.t. $mIoU$. On average over all scenes, we achieve better performance with $0.812$ compared to $0.781$ of the state-of-the-art method 4DMOS.

4DMOS considerably outperforms our approach on the ``Campus 1'' scene. As shown in Figure~\ref{fig:qualitative_campus}, our approach makes wrong segmentation predictions on the ground, some occluded parts while an object passes by, and not the entire vehicle is covered. 4DMOS achieves an almost perfect overlap with the ground truth. At the same time, LMNet misses out on some pedestrians. In contrast, our approach accurately segments the pedestrians. 

Our unsupervised approach outperforms the supervised approaches on the ``City 1'' and ``City 2'' scenes. Based on the visualization in Figure~\ref{fig:qualitative_city}, we now pose possible explanations for the lower performance of LMNet and 4DMOS.
Especially for LMNet, the drop in performance is due to (a)~wrongly segmenting vehicles that are standing still next to the ego-vehicle and (b)~missing some pedestrians that are far away. We hypothesize that LMNet encounters an out-of-distribution scenario. Stopping at a red light with cars around the ego-vehicle may not appear often enough in the training data. However, the training data includes various highway/secondary road scenes with cars close to the ego-vehicle that keep the same distance by driving at a similar speed. In such scenarios, cars close to the ego-vehicle also move, and LMNet can segment them as moving objects. When the ego-vehicle is stationary, the model seems to not generalize well. Therefore, our experiment indicates biased training data w.r.t. non-stationary scenes in SemanticKITTI, which may cause the performance to drop for LMNet in this stationary setting. In contrast, 4DMOS shows excellent generalization capability to the stationary setting, but it misses a moving vehicle in the upper part of the scene, which our approach segments correctly.




\subsection{Influence of Hyperparameters}

\begin{figure*}
    \centering
    \includegraphics[width=0.9\linewidth]{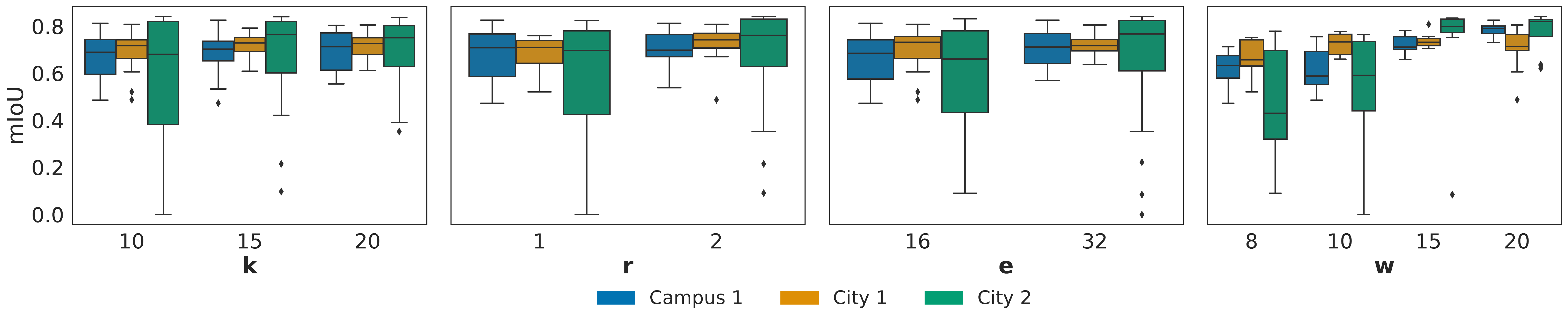}    \caption{Ablation study regarding number of clusters $k$, radius $r$, embedding dimension $e$, and window size $w$}
    \label{fig:ablation}
\end{figure*}

\begin{table}[t]
    \centering
    
    \begin{tabular}{cc}
    \toprule
    Approach Configuration $k\in\{10,15,20\}$ & $mIoU (\uparrow)$\\
     \midrule     
    \textbf{$r=1, e=16, w=15$}  & $.756 \pm .025$   \\
     \textbf{$r=1, e=16, w=20$}  & $.757 \pm .066$    \\
    \textbf{$r=2, e=32, w=15$}  & $ .759 \pm .055 $    \\
      \textbf{$r=1, e=32, w=20$}  & $.774 \pm .052$   \\
      \textbf{$r=2, e=32, w=20$}  & $.800 \pm .043$    \\
        \bottomrule
    \end{tabular}
    \caption{Top five average $mIoU$ results over all clusters and scenes}
    \label{tab:cluster_averaged}
\end{table}


In this section, we evaluate the influence of hyperparameters $e, r, k,$ and $w$ on the performance of our approach. We conducted an experiment with different configurations. The results are summarized in Figure~\ref{fig:ablation} and Table~\ref{tab:cluster_averaged}. 

In Figure~\ref{fig:ablation}, the respective x-axis of each subplot varies different settings for (a) the number of clusters~$k$, (b) the size of radius~$r$, (c) the embedding dimension~$e$, and (d) the window size~$w$. 
The y-axis shows the averaged $mIoU$ performance of different configurations over the three scenes in dependency to the varied parameter on the x-axis. We present the results using a boxplot to visualize the standard deviation across different configurations, which serves as an uncertainty measure.
Table~\ref{tab:cluster_averaged} summarizes the five best configurations w.r.t. $r$, $e$, and $w$ across different cluster numbers and all scenes. We computed the mean over the respective $mIoU$ results. 

\textbf{Impact of number of clusters.} Figure~\ref{fig:ablation} shows how the number of clusters influences the performance of our approach. Our approach reaches good performance more consistently with 15 and 20 clusters across different parameter configurations. We attribute this result to different patterns of the stationary or moving parts (\eg, corners, walls, ground, pillars, trees) in the scene that are captured by MOTS. Hence, more clusters are needed to correctly partition the latent space. 

\textbf{Impact of radius.}
Regarding the radius $r$, Figure~\ref{fig:ablation} shows that on average a higher radius yields a better $mIoU$. A higher radius implies a larger receptive field, which benefits the encoder to distinguish between moving and stationary patterns in MOTS. More specifically, the averaged results over all clusters and scenes in Table~\ref{tab:cluster_averaged} show that our approach reaches best performance with a radius of $r=2$. 

\textbf{Impact of embedding dimension.}
The best performance across all cluster configurations is achieved with $e=32$ as shown in Table~\ref{tab:cluster_averaged}. In fact, the three best configurations all used $e=32$, which suggests that a higher embedding dimension achieves the best performance. However, we cannot conclude from our study in Figure~\ref{fig:ablation} that embedding dimension $e=32$ consistently outperforms $e=16$.  

\textbf{Impact of window size.}
The top five results in Table~\ref{tab:cluster_averaged} show that larger window sizes from $w=15$ to $w=20$ perform the best overall. In Figure~\ref{fig:ablation}, we can observe that both $w=15$ and $w=20$ have considerably better performance across all scenes. Our experiments show that context concerning the 15--20 past frames is the most effective configuration for stationary scenes. Twenty past frames correspond to two seconds temporal context at a 10Hz framerate.



\subsection{Qualitative Results with a Velodyne VLP-16}
We trained one model with the best parameter configuration ($r=2, e=32$) on the first 200 frames of seven stationary scenes for 5 epochs. Furthermore, we use $k=20$ clusters when training the GMM. 
Because the VLP-16 data was recorded at 20Hz, we used a window size of $w=40$ to match the best performing temporal history of 2 seconds. This shows that our approach can even scale to temporal histories greater than 20 frames. In contrast, other approaches (\eg, 4DMOS) may not be able to handle such a long history due to the enormous memory consumption. 

The qualitative results on a leave-out test scene in Figure~\ref{fig:qualitative_vlp16} show that our approach can be applied to different LiDAR sensor setups (\eg, VLP-16, HDL-64E) that even have different temporal resolutions (\eg, 10Hz, 20Hz). We can see that our approach accurately segments movement of different pedestrians and a cyclist. Wrong predictions are obtained for tree leaves due to noisy sensor measurements that probably appear similar to movement in MOTS.

\section{Discussion and Future Work}

Our experimental evaluation shows the potential of our approach to segment moving objects in stationary LiDAR point cloud videos without any supervision.
The proposed approach for stationary LiDAR sensors can outperform state-of-the-art supervised models as shown in Table~\ref{tab:miou_raw_kitti}. 
However, we want to emphasize that our model has limitations in a non-stationary setting. In particular, points entering the scene or previously occluded parts that become visible appear exactly the same in MOTS when compared to a moving object. Extending the approach for non-stationary LiDAR data is left for future work.

Furthermore, our approach is limited by a small receptive field. For instance, recent works on Vision Transformers show that global context is essential for learning good feature representations~\cite{dosovitskiy2020image, yang2021focal, Cho_2021_CVPR}. Increasing the MOTS receptive field by a small amount implies a quadratic scaling of MOTS channels, which quickly scales to thousands of channels. The latter results in strong performance limitations w.r.t. time and memory, because one MOTS is computed for each unique voxel in the scene for each frame. For this reason, future work on our approach includes finding an efficient method for scaling the receptive field.



\section{Negative Sociental Impact}
Smart cities of the future will have an infrastructure to collect enormous amounts of data from heterogeneous data sensors such as LiDAR, surveillance cameras, or temperature sensors. These sensors build the foundation for a digital twin that can reason about all kinds of behavior in the city~\cite{shahat2021city}. This information will help to enhance the lives of citizens and have a substantial impact on intelligent transportation systems~\cite{harris2019mlk} in the future. However, using surveillance cameras as a data source for digital twins raises strong privacy concerns. For instance, cameras capture color information of the natural scene, allowing person re-identification~\cite{dietlmeier2021important}. In the wrong hands, this information encourages tracking a specific target, potentially leading to blackmail, which results in a strong negative sociental impact. For this reason, we want to raise awareness on using LiDAR as a more anonymity-preserving technology for surveillance. That is because LiDAR does not record facial characteristics or further details like hair and skin color~\cite{velodyne2020lidar}. In a stationary setting, LiDAR sensors can detect all kinds of objects and reason about their behavior and may replace the need for RGB cameras in public places. 

\section{Conclusion}

This work addresses unsupervised moving object segmentation (MOS) in stationary LiDAR point cloud videos. Our approach effectively learns voxel embeddings from occupancy changes in a spatio-temporal neighborhood. We propose to model the occupancy changes in the neighborhood of a voxel by a multivariate occupancy time series (MOTS), which in turn allows learning voxel embeddings that encode motion information. As a consequence, our MOTS voxel representation relaxes unsupervised MOS to a multivariate time series clustering problem. We evaluate our method quantitatively on stationary scenes from the Raw KITTI dataset and qualitatively on stationary VLP-16 data. We achieve comparable performance to state-of-the-art supervised MOS approaches in the stationary setting.


\section*{Acknowledgement}
This work has been funded by the LOEWE initiative (Hesse, Germany) within the emergenCITY centre.

{\small
\bibliographystyle{ieee_fullname}
\bibliography{egbib}
}

\end{document}